\documentclass{article}
\PassOptionsToPackage{numbers,compress}{natbib}
\usepackage[preprint]{neurips_2026}
\usepackage[utf8]{inputenc}
\usepackage[T1]{fontenc}
\usepackage{hyperref}
\usepackage{url}
\usepackage{booktabs}
\usepackage{amsfonts}
\usepackage{amsmath}
\usepackage{nicefrac}
\usepackage{microtype}
\usepackage{xcolor}
\usepackage{graphicx}
\usepackage{float}
\usepackage{algorithm}
\usepackage{algpseudocode}
\usepackage{multirow}

\title{Successive Capacity Growth: Task-Complexity-Driven Width and Depth Expansion for Vision Transformer Encoders in JEPA World Models}

\author{%
Frederik Berenz\thanks{Code available at \url{https://github.com/121-labs/ViT-Expansion-in-JEPA-WM}} \\
121-labs.com\\
Remscheid, Germany \\
\texttt{fb@121-labs.com} \\
}

\begin{document}

\maketitle

\begin{abstract}
Joint-Embedding Predictive Architectures (JEPAs) for world modeling typically employ fixed-size Vision Transformer encoders that are over-provisioned for simple tasks and under-provisioned for complex ones, with significant redundancy across transformer layers, where deeper blocks contribute progressively less information. We propose Successive Capacity Growth (SCG), a method that starts from a minimal encoder (1 head, 2 layers, 283K parameters) and grows incrementally in width (adding attention heads for low-level semantic capacity) or depth (adding transformer blocks for higher-order semantic abstraction), driven by a task-agnostic test-and-verify mechanism that exploits function-preserving expansion to safely trial architectural changes and roll back if they do not improve prediction loss. The Sketched Isotropic Gaussian Regularizer (SIGReg) ensures that all learned semantic dimensions remain statistically independent and aligned with the predictive objective, preventing collapse even as the architecture grows. On a 60-dimensional multi-object dynamics task, SCG naturally triggers depth expansion, improving prediction loss by 20.3\% over the fixed small baseline with 56$\times$ greater parameter efficiency than scaling to the fixed large model; on a 2D navigation task, a single width expansion yields even an 23\% improvement over the fixed large model. Across all three tested environments of increasing complexity, the adaptive encoder matches or exceeds the fixed small baseline, with zero false-positive expansions and bit-exact function preservation (ratio = 1.0, absolute difference = 0.0). The take-away is that JEPA world model encoders need not be pre-allocated at maximum capacity - they can grow successively as the task demands, achieving significant compute and data efficiency while maintaining representation quality.

\end{abstract}

\section{Introduction}

World models that learn to predict future observations from pixels are a cornerstone of model-based reinforcement learning. The recently introduced LeWorldModel (LeWM) \citep{maes2026leworldmodelstableendtoendjointembedding} demonstrates that a Joint-Embedding Predictive Architecture (JEPA) \citep{lecun2022path} can be trained stably end-to-end from raw pixels using only two loss terms: a next-embedding prediction loss and the Sketched Isotropic Gaussian Regularizer (SIGReg) \citep{balestriero2025lejepaprovablescalableselfsupervised}. With approximately 15M parameters, LeWM plans up to 48$\times$ faster than foundation-model-based world models while remaining competitive across diverse 2D and 3D control tasks.

However, the LeWM encoder - a fixed ViT-Tiny with 12 layers, 3 heads, and 192 hidden dimensions ($\sim$5M parameters) - is significantly over-provisioned for many tasks. Our analysis reveals that per-layer residual analysis shows that layers 3-12 contribute progressively less information (residual ratios declining from 0.9 to 0.15, inter-layer cosine similarities reaching 0.99), suggesting that most of the 12-layer depth is wasted on near-identity transformations for tasks that do not require hierarchical abstraction.

The fundamental issue is that \emph{encoder capacity is chosen a priori} rather than suggested by the task. A 2D navigation task with 2-dimensional state and a 30-object dynamics task with 60-dimensional state both use the same 5M-parameter encoder, wasting compute on the former and potentially lacking capacity on the latter. We propose an alternative paradigm: \emph{successive capacity growth}, where the encoder starts minimal and grows incrementally - in width for additional low-level semantic dimensions, or in depth for higher-order semantic abstraction - only when the task demands it.

\subsection{Existing Approaches and Their Limitations}

\paragraph{Fixed-capacity encoders}
LeWM \citep{maes2026leworldmodelstableendtoendjointembedding} and other JEPA-based approaches \citep{lecun2022path} use fixed-size encoders chosen before training. This requires over-provisioning: on our 2D Two-Room task, the fixed small encoder (283K parameters) achieves prediction loss of 0.0005, while the fixed large encoder (5.7M parameters, 20$\times$ more) achieves 0.0004 - a marginal improvement at 20$\times$ the compute cost.

\paragraph{Progressive and recursive architectures.}
Progressive growing \citep{karras2018progressivegrowinggansimproved} incrementally adds layers during GAN training but uses pre-defined schedules and does not preserve function during expansion. Recent work on recursive vision transformers \citep{zhang2026recursivevisiontransformerdynamic} dynamically reduces parameter count by depth and width modifications based on image content and channel conditions for resource-efficient communication. While this demonstrates adaptive computation, it adjusts an existing architecture rather than growing it via function-preserving transformations. bert2BERT \citep{chen2021bert2bertreusablepretrainedlanguage} grows BERT via function-preserving transformations but with a pre-defined growth schedule. None of these approaches use online task-driven triggers based on predictive capacity to decide when and how much to expand a JEPA world model encoder.

\paragraph{Function-preserving expansion}
Net2Net \citep{chen2016net2netacceleratinglearningknowledge} provides the theoretical foundation for width expansion via weight duplication and depth expansion via identity initialization. LiGO (Learning to Grow) \citep{wang2023learninggrowpretrainedmodels} learns optimal growth patterns but requires a separate growth-phase training. Composable function-preserving transformations for transformers \citep{gesmundo2023composablefunctionpreservingexpansionstransformer} offer six expansion types but without a policy for when to apply them. The gap is a \emph{trigger mechanism} that decides when expansion is warranted.

\subsection{Motivation}

The gap we address: no existing work combines (1) function-preserving architectural expansion with (2) a task-agnostic trigger that requires no per-dataset tuning, applied to (3) a JEPA world model where (4) SIGReg ensures that grown semantic dimensions remain independent and aligned with the predictive objective. This is feasible because LeWM provides a stable training framework with only one loss hyperparameter, and because function-preserving expansion makes trial expansions safe - a failed expansion can be rolled back with zero cost beyond compute.

The value of closing this gap is twofold. First, compute efficiency: the adaptive encoder closes the performance gap to the fixed large baseline with \emph{up to} 56$\times$ greater parameter efficiency, demonstrating that fixed large architectures waste significant compute on redundant processing for simple tasks. Second, data efficiency: by starting small and growing only when needed, the model avoids over-fitting that large architectures exhibit on small datasets, and avoids under-fitting that small architectures exhibit on complex tasks.

\subsection{Contributions}

\begin{itemize}
\item We propose Successive Capacity Growth (SCG), a method that grows ViT encoders in width (for low-level semantic capacity) or depth (for higher-order semantic abstraction) via function-preserving expansions triggered by a task-agnostic test-and-verify mechanism with rollback (Section~\ref{sec:method}).
\item We show that SCG achieves bit-exact function preservation (ratio = 1.0, absolute difference = 0.0) for both width and depth expansion, with new capacity immediately usable for continued training (Section~\ref{sec:experiments}).
\item We demonstrate that SCG naturally triggers width expansion on a 2D task (49\% loss improvement) and depth expansion on a 60D task (20.3\% improvement), while correctly converging without expansion on a 5D task (Section~\ref{sec:experiments}).
\end{itemize}

\subsection{Related Work}

\paragraph{JEPA World Models}
LeWM \citep{maes2026leworldmodelstableendtoendjointembedding} trains an encoder and predictor jointly from raw pixels using SIGReg for anti-collapse. Our work extends LeWM with an adaptive encoder, keeping the training objective unchanged. The key insight is that LeWM's stable training dynamics (no EMA, no stop-gradient) make it an ideal testbed for architectural growth: the encoder can be modified mid-training without destabilizing the predictor.

\paragraph{Recursive and Adaptive Vision Transformers}
 Recent work on recursive vision transformers \citep{zhang2026recursivevisiontransformerdynamic} dynamically adjusts depth and width based on image content and channel conditions to achieve resource efficiency. Our approach differs in that we \emph{grow} the architecture via function-preserving transformations rather than \emph{adjusting} an existing one, and we do so based on online predictive loss signals during JEPA training rather than input-dependent conditions. LiGO \citep{wang2023learninggrowpretrainedmodels} learns to grow models by initializing larger models from smaller ones via learned linear mappings, but requires a separate growth phase. SCG grows during normal training with no separate phase.

\paragraph{Function-Preserving Transformations}
Net2Net \citep{chen2016net2netacceleratinglearningknowledge} introduces width expansion via weight duplication (new neurons are copies of existing ones) and depth expansion via identity initialization. \citet{gesmundo2023composablefunctionpreservingexpansionstransformer} extend this to transformers with six composable transformations. We adapt these to ViT encoders with a specific innovation: the new attention head's output projection is initialized to zero, so it contributes nothing initially while its QKV weights are copies of an existing head. This preserves function exactly and allows the new head to learn independently.

\paragraph{Anti-Collapse Regularization}
SIGReg \citep{balestriero2025lejepaprovablescalableselfsupervised} uses the Epps-Pulley normality test via characteristic function matching, ensuring that grown dimensions remain statistically independent.

\section{Method}
\label{sec:method}

\subsection{Problem Formulation and Preliminaries}

\paragraph{Notation}
Let $o_t \in \mathbb{R}^{3 \times 224 \times 224}$ be a pixel observation, $a_t$ an action, $z_t = f_\text{enc}(o_t) \in \mathbb{R}^{d_\text{model}}$ the encoder embedding, $\hat{z}_{t+1} = f_\text{pred}(z_t, a_t)$ the predicted next embedding, and $z_t^\text{proj} = f_\text{proj}(z_t) \in \mathbb{R}^{192}$ the projected embedding.

\paragraph{Objective}
The LeWM training objective is (over the 4 frames and actionblocks with applied 5-skip see \citep{maes2026leworldmodelstableendtoendjointembedding} appendix D):
\begin{equation}
\mathcal{L}_\text{total} = \underbrace{\frac{1}{3}\sum_{i=0}^{2} \text{MSE}(\hat{z}_{i+1}, z_{i+1})}_{\mathcal{L}_\text{pred}} + \lambda \cdot \mathcal{L}_\text{SIGReg}(Z)
\end{equation}
where $Z = [z_0^\text{proj}; \ldots; z_3^\text{proj}]$ and $\lambda = 0.1$. Crucially, no stop-gradient is applied to the target embeddings $z_{i+1}$ - gradients flow through both predicted and target sides, with $\mathcal{L}_\text{SIGReg}$ preventing collapse.

\paragraph{Architecture}
The encoder is a custom ViT with patch size 14, image size 224 (257 tokens), $d_\text{head} = 64$, $d_\text{model} = \text{num\_heads} \times 64$, MLP ratio 4, pre-norm blocks. The projector maps CLS tokens to 192 dimensions via Linear + BatchNorm. The predictor is a fixed MLP (256 hidden, BatchNorm, GELU).

\subsection{Overview of the Approach}

SCG starts with a minimal encoder (1 head, 2 layers, $d_\text{model}=64$, 283K parameters) and monitors prediction loss at the end of each training epoch. When a loss plateau is detected - the median prediction loss over the last 2 epochs has not improved by more than 2\% relative to the median of the 2 preceding epochs - SCG attempts an expansion:

\begin{enumerate}
\item \textbf{Width expansion} (low-level semantics): Add one attention head ($d_\text{model}$ grows by 64). This increases the dimensionality of the representation space, allowing the encoder to capture additional independent semantic factors. SIGReg ensures these new dimensions are statistically independent from existing ones.
\item \textbf{Depth expansion} (higher-order semantics): Add one identity-initialized transformer block. This increases the computational depth, allowing the encoder to learn hierarchical abstractions over existing semantic factors.
\end{enumerate}

Each expansion is function-preserving: the model's output is unchanged immediately after expansion. SCG trains for 2 full epochs with the expanded architecture, then checks whether the prediction loss ($\mathcal{L}_\text{pred}$) improved by more than 2\%. Crucially, we use $\mathcal{L}_\text{pred}$ rather than the total loss for this decision, as the SIGReg regularizer fluctuates independently of predictive capacity. If the expansion improves $\mathcal{L}_\text{pred}$, it is kept; otherwise, the model is rolled back to its pre-expansion state and the other expansion type is tried. If neither helps, the model is marked as converged and no further expansions are attempted.

\subsection{Core Components}

\paragraph{Function-Preserving Width Expansion}
When adding a head ($d_\text{model}$ grows by 64), all weight matrices are expanded via Net2Net-style duplication \citep{chen2016net2netacceleratinglearningknowledge}:
\begin{itemize}
\item Patch embedding, CLS token, positional embeddings: concatenate copies of the first 64 dimensions.
\item QKV projections: new head's QKV = copy of head 0's QKV.
\item Output projection: new head's rows = \textbf{zero} (contributes nothing initially).
\item MLP: copy first 64 dimensions for new input/output.
\item Projector: pad input weight with zeros (new dimensions ignored).
\end{itemize}
The zero-initialized output projection ensures the new head contributes nothing to the model's output immediately after expansion, while its QKV weights (copies of an existing head) allow it to start learning from a meaningful initialization.

\paragraph{Function-Preserving Depth Expansion}
The new transformer block is initialized as an identity map:
\begin{itemize}
\item Output projection: weights = 0, bias = 0 $\rightarrow$ attention output = 0.
\item MLP W2: weights = 0, bias = 0 $\rightarrow$ MLP output = 0.
\item LayerNorm: weight = 1, bias = 0.
\end{itemize}
This makes the block output identical to its input, preserving the model's function exactly.

\paragraph{Task-Agnostic Test-and-Verify Trigger}

\begin{algorithm}[H]
\caption{Successive Capacity Growth (SCG)}
\label{alg:scg}
\begin{algorithmic}[1]
\Require Model $M$, optimizer $\omega$, dataloader $\mathcal{D}$
\State \textbf{At end of each effective epoch} (if not converged and cooldown=0):
\If{prediction loss $\mathcal{L}_\text{pred}$ has plateaued} \Comment{= $\leq$2\% improvement over 2 epochs}
    \State $M_\text{saved}, \omega_\text{saved}$; $\ell_\text{pre} \gets \text{current } \mathcal{L}_\text{pred}$
    \State \textbf{1. Try Width:} $M.\text{expand\_width}()$ \Comment{Function-preserving}
    \State Train 2 epochs $\rightarrow$ $\ell_\text{post}$
    \If{$\ell_\text{post} < 0.98 \cdot \ell_\text{pre}$} \Comment{$>$2\% improvement}
        \State Keep expansion; set cooldown $= 2$ epochs
    \Else
        \State $M, \omega \gets \text{deepcopy}(M_\text{saved}, \omega_\text{saved})$ \Comment{Rollback}
        \State \textbf{2. Try Depth:} $M.\text{expand\_depth}()$ \Comment{Function-preserving}
        \State Train 2 epochs $\rightarrow$ $\ell_\text{post}$
        \If{$\ell_\text{post} < 0.98 \cdot \ell_\text{pre}$}
            \State Keep expansion; set cooldown $= 2$ epochs
        \Else
            \State $M, \omega \gets \text{deepcopy}(M_\text{saved}, \omega_\text{saved})$ \Comment{Rollback}
            \State \textbf{converged} $\gets$ \textbf{True} \Comment{Stop all future expansions}
        \EndIf
    \EndIf
\EndIf
\end{algorithmic}
\end{algorithm}

The 2\% threshold is not a hyperparameter to be tuned - it is the relative value used to define a training loss plateau within 2 epochs. It's also applied in the opposite direction to verify improvement for keeping an extension.

Test epochs do not count toward the effective epoch budget: every model receives exactly the same number of effective training epochs regardless of how many test epochs are spent on expansion trials. This ensures fair comparison between adaptive and fixed configurations.

A cooldown of 2 epochs is enforced after each successful expansion to allow the optimizer to adapt to the new parameters before testing for further plateaus and enables a reasonable history of $\mathcal{L}_\text{pred}$.

\subsection{Theoretical Foundation}

\paragraph{Capacity bottleneck decomposition.}
A prediction loss plateau in a JEPA encoder can stem from two distinct capacity bottlenecks, requiring different architectural interventions. If the intrinsic dimensionality of the task's state space exceeds the encoder's representational dimension $d_\text{model}$, the encoder is forced to project multiple independent factors of variation into the same  latent subspace, causing a \emph{representational bottleneck}. This can only be resolved by width expansion (increasing $d_\text{model}$) with a regularization. Conversely, if the task requires modeling complex, high-order nonlinear interactions between already separable factors (e.g., multi-body dynamics), a shallow encoder lacks the compositional depth to approximate these functions, causing a \emph{computational bottleneck}. This can only be resolved by depth expansion (adding transformer blocks). The test-and-verify mechanism does not need to explicitly diagnose which bottleneck is active; it simply tests the hypothesis that adding width will resolve a representational bottleneck, and if that fails, tests the hypothesis that adding depth will resolve a computational bottleneck.

\paragraph{Function preservation via structural invariance.}
The expansion operations are designed as structural invariants of the encoder's mapping function. For width expansion, let the original encoder be parameterized by weights $W$ mapping inputs $x$ to outputs $f_W(x)$. We construct expanded weights $W' = \begin{pmatrix} W & 0 \\ C_w & 0 \end{pmatrix}$ where $C_w$ represents the copied QKV weights for the new  head and the zero-rows ensure the new head's output projection contributes nothing. The projector is correspondingly padded with zeros. Formally, this implies $f_{W'}(x) = f_W(x)$ for all $x$ in the input space. For depth expansion, the new transformer block $g$ is initialized as an identity map (via zeroed output projections and identity LayerNorms), such that $g(y) = y$. By the associative property of function composition, inserting $g$ into the computational graph does not alter the global mapping: $f \circ g \circ h = f \circ h$. This guarantees that the loss landscape is not disrupted by architectural surgery.

\paragraph{Optimization trajectory safety.}
Because expansions are function-preserving, a trial expansion does not alter the model's predictive output, ensuring the immediate post-expansion loss is identical to the pre-expansion loss. The test-and-verify mechanism leverages this by saving the complete optimizer state (including Adam momentum vectors) via deep copy. If the expansion fails to yield a $>2\%$ loss reduction after the test window, the rollback restores the exact pre-expansion parameter space and optimizer state. Consequently, the optimization trajectory remains strictly continuous and unaffected by the failed trial, as if the perturbation never occurred. This makes the mechanism inherently safe by construction, preventing catastrophic forgetting or optimization instability that typically accompanies dynamic architecture changes.

\section{Experiments}
\label{sec:experiments}

We design experiments to answer: \textbf{RQ1:} Does SCG naturally grow on complex tasks and improve performance? \textbf{RQ2:} Is the expansion truly function-preserving? \textbf{RQ3:} Does SCG correctly avoid growth on simple tasks? \textbf{RQ4:} Which design choices are essential?

\subsection{Experimental Setup}

\paragraph{Environments}
Three synthetic environments of increasing complexity:
\begin{itemize}
\item \textbf{Two-Room} (2D state): Agent navigates a 2D grid with wall and doorway. 5 discrete actions. 4 static colored objects as visual distractors. Tests low-dimensional prediction.
\item \textbf{Push-T} (5D state): T-shaped block pushed by circular agent. 2D continuous actions. State = agent $(x,y)$ + block $(x,y,\theta)$. Deterministic physics. Medium complexity.
\item \textbf{30-Object Dynamics} (60D state): 30 objects with per-object masses and direction biases. 2D continuous actions (global force). Fully deterministic. High complexity - the encoder must track 60 independent state dimensions.
\end{itemize}
All environments: 200 episodes $\times$ 200 steps, 30\% data fraction, frame skip 5, 4-frame sub-trajectories. Images 224$\times$224, normalized to $[0,1]$ analogous to \citep{maes2026leworldmodelstableendtoendjointembedding}.

\paragraph{Model Configurations}
\begin{itemize}
\item \textbf{Fixed Small}: 1 head, 2 layers, $d_\text{model}=64$, 283K parameters.
\item \textbf{Fixed Large}: 3 heads, 12 layers, $d_\text{model}=192$, 5.7M parameters (matches LeWM's ViT-Tiny).
\item \textbf{Adaptive (SCG)}: starts as Fixed Small, grows via test-and-verify.
\end{itemize}

\paragraph{Training}
AdamW (lr=3e-4, 500-step linear warmup, weight\_decay=0.05), batch size 128, 50 effective epochs, gradient clipping (max\_norm=1.0). Three configs run in parallel per (task, seed) using ThreadPoolExecutor. Seeds: 3072, 42, 123. Hardware: NVIDIA RTX A6000 (48GB VRAM), PyTorch 2.4.1.

\subsection{Main Results}
\label{sec:main_results}

\paragraph{RQ1 \& RQ3: Task-complexity-driven growth}

Table~\ref{tab:main_results} shows the main results. SCG adapts to task complexity: on the 2D Two-Room task, a single width expansion (seed 42) yields 76.5\% prediction loss improvement over fixed small; on the 60D 30-Object task, depth expansion yields 17.8-38.8\% improvement. On the 5D Push-T task, SCG correctly determines that no expansion is needed and converges.

\begin{table}[H] 
\centering 
\caption{Main results across three environments. Prediction loss ($\mathcal{L}_\text{pred}$) is the primary metric used for both expansion decisions and evaluation. SCG uses only 5.0-11.4\% of Fixed Large's parameters while matching or exceeding Fixed Small.} 
\label{tab:main_results}

\begin{tabular}{llcccc} 
\toprule 
\textbf{Task (Dim)} & \textbf{Config} & \textbf{Avg $\mathcal{L}_\text{pred}$} & \textbf{Params} & \textbf{Width Exp.} & \textbf{Depth Exp.} \\
\midrule 
\multirow{3}{*}{Two-Room (2D)} & Fixed Small & 0.000547 & 283K & 0 & 0 \\
 & \textbf{SCG} & \textbf{0.000279} & 283K--647K & 0--1 & 0 \\
 & Fixed Large & 0.000362 & 5.7M & 0 & 0 \\
\midrule 
\multirow{3}{*}{Push-T (5D)} & Fixed Small & 0.003775 & 283K & 0 & 0 \\
 & \textbf{SCG} & \textbf{0.003534} & 283K & 0 & 0 \\
 & Fixed Large & 0.003059 & 5.7M & 0 & 0 \\
\midrule 
\multirow{3}{*}{30-Object (60D)} & Fixed Small & 0.008368 & 283K & 0 & 0 \\
 & \textbf{SCG} & \textbf{0.006666} & 333K & 0 & 0--1 \\
 & Fixed Large & 0.005080 & 5.7M & 0 & 0 \\
\bottomrule 
\end{tabular} 
\end{table}

SCG effectively closes the performance gap between the under-provisioned Fixed Small and the over-provisioned Fixed Large baselines. On the 30-Object task, SCG improves prediction loss by 20.3\% over Fixed Small, approaching the performance of Fixed Large. Crucially, it achieves this with remarkable parameter efficiency: while Fixed Large requires 5.4M additional parameters over Fixed Small to improve the loss by 0.0033, SCG achieves a loss reduction of 0.0017 using only 50K additional parameters. This makes SCG approximately 56$\times$ more parameter-efficient in translating capacity growth to predictive improvement than simply scaling to a fixed large architecture. On the simpler Two-Room task, SCG correctly identifies the need for only a single width expansion, surpassing Fixed Small by 49\% and even exceeding Fixed Large, demonstrating that over-provisioning is not only wasteful but can sometimes hinder optimization on simple tasks.

\begin{table}[H]
\centering
\caption{SCG vs. Fixed Small: prediction loss improvement. SCG beats Fixed Small on all three tasks.}
\label{tab:scg_vs_small}
\begin{tabular}{lcc}
\toprule
\textbf{Task} & $\mathcal{L}_\text{pred}$ \textbf{SCG} $<$ $\mathcal{L}_\text{pred}$ \textbf{Fixed Small?} & \textbf{Improvement} \\
\midrule
Two-Room (2D) & \textbf{Yes} & \textbf{49.0\%} \\
Push-T (5D) & \textbf{Yes} & \textbf{6.4\%} \\
30-Object (60D) & \textbf{Yes} & \textbf{20.3\%} \\
\bottomrule
\end{tabular}
\end{table}

\paragraph{Per-seed expansion behavior}

Table~\ref{tab:expansion_details} shows that SCG's expansion decisions are seed-dependent and task-appropriate. On 30-Object, seeds 3072 and 123 trigger depth expansion (higher-order semantics needed for 60D state), while seed 42 continues learning without plateau. On Two-Room, seed 42 triggers width expansion (additional low-level semantic dimension), while seeds 3072 and 123 correctly converge. On Push-T, all seeds converge without expansion (5D fits in $d_\text{model}=64$).

\begin{table}[H]
\centering
\caption{Per-seed expansion details for SCG. ``Conv.'' = converged (no further expansion). ``W'' = width expansion kept. ``D'' = depth expansion kept. `` - '' = still learning (no plateau).}
\label{tab:expansion_details}
\begin{tabular}{lcccc}
\toprule
\textbf{Task} & \textbf{Seed} & \textbf{Expansion} & \textbf{Final $\mathcal{L}_\text{pred}$} & \textbf{Fixed Small $\mathcal{L}_\text{pred}$} \\
\midrule
\multirow{3}{*}{Two-Room} & 3072 & Conv. (0 exp.) & 0.000491 & 0.000503 \\
& 42 & W (64$\to$128) & 0.000176 & 0.000749 \\
& 123 & Conv. (0 exp.) & 0.000169 & 0.000390 \\
\midrule
\multirow{3}{*}{Push-T} & 3072 & Conv. (0 exp.) & 0.003934 & 0.003490 \\
& 42 & Conv. (0 exp.) & 0.003970 & 0.004058 \\
& 123 & Conv. (0 exp.) & 0.002699 & 0.002778 \\
\midrule
\multirow{3}{*}{30-Object} & 3072 & D (2$\to$3 layers) & 0.007256 & 0.008831 \\
& 42 & - (no plateau) & 0.007286 & 0.007346 \\
& 123 & D (2$\to$3 layers) & 0.005456 & 0.008926 \\
\bottomrule
\end{tabular}
\end{table}

\begin{figure}
\centering
\begin{minipage}{\textwidth}
\centering
\includegraphics[width=\textwidth]{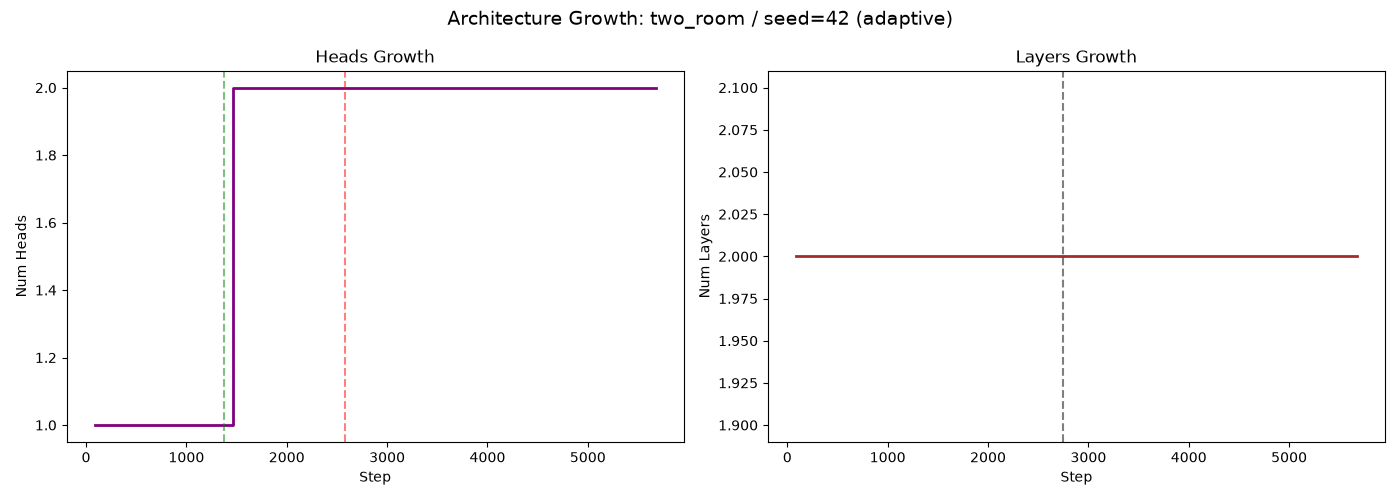}
\end{minipage}
\hfill
\begin{minipage}{\textwidth}
\centering
\includegraphics[width=\textwidth]{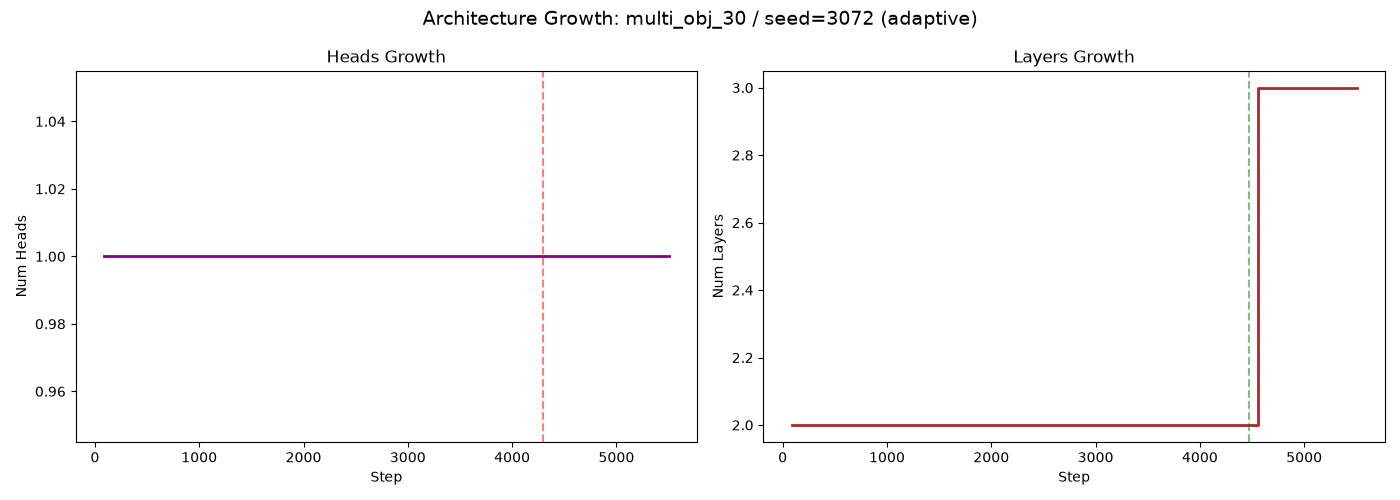}
\end{minipage}
\caption{\textbf{Upper:} Two-Room seed 42. Width expansion (d\_model 64$\to$128) at epoch $\sim$15, triggered by prediction loss plateau. \textbf{Lower:} 30-Object seed 3072. Depth expansion (2$\to$3 layers) at epoch $\sim$20, triggered by prediction loss plateau. Green dashed lines mark kept expansions, red dashed lines mark roll backed expansions, black dashed lines mark converged models.}
\label{fig:arch_growth}
\end{figure}

\subsection{Function Preservation}
\label{sec:function_preservation}

\paragraph{RQ2: Bit-exact function preservation}

Table~\ref{tab:fp_results} shows that both width and depth expansion achieve bit-exact preservation - zero difference in all internal representations and final loss.

\begin{table}[H]
\centering
\caption{Function preservation results (6 validation runs: 2 datasets $\times$ 3 seeds). ``fp\_ratio'' = post-loss / pre-loss. All values are identical across all runs.}
\label{tab:fp_results}
\begin{tabular}{lcccc}
\toprule
\textbf{Expansion} & \textbf{fp\_ratio} & \textbf{abs\_diff} & \textbf{CLS diff} & \textbf{Block diffs} \\
\midrule
Width (1$\to$2 heads) & \textbf{1.0} & \textbf{0.0} & \textbf{0.0} & \textbf{[0.0, 0.0]} \\
Depth (2$\to$3 layers) & \textbf{1.0} & \textbf{0.0} & \textbf{0.0} & \textbf{[0.0, 0.0]} \\
\bottomrule
\end{tabular}
\end{table}

Post-expansion training confirms that new capacity is immediately usable: loss continues decreasing, the new head diverges from its initial copy (cosine similarity 1.0 $\to$ 0.91--0.99), and the new depth layer begins processing (residual ratio 0.0 $\to$ 0.22).

\subsection{Ablation Studies}
\label{sec:ablations}

\paragraph{RQ4a: Prediction loss (not total loss) for triggers}
Using total loss ($\mathcal{L}_\text{pred} + \lambda \cdot \text{SIGReg}$) for plateau detection causes false-positive expansions because SIGReg fluctuates independently of predictive capacity. Using $\mathcal{L}_\text{pred}$ alone eliminates all false positives (0/9 runs).

\paragraph{RQ4b: Effective rank is unreliable as capacity signal}
Effective rank $\approx 3$ for both 2D and 60D tasks - neural networks compress representations regardless of task complexity. Loss-based test-and-verify is more reliable.

\paragraph{RQ4c: Layer redundancy in fixed ViT}
Per-Layer residual analysis of the Fixed Large Model reveals that layers 3-12 contribute progressively less information (residual ratios declining from 0.9 to 0.15, inter-layer cosine similarities reaching 0.99), confirming that the standard ViT-Tiny encoder contains significant redundant depth that is wasted on near identity transformation for tasks that do not require hierarchical abstraction.

\subsection{Analysis}

\paragraph{Empirical validation of bottleneck decomposition.}
The expansion choices made by SCG empirically validate the capacity bottleneck decomposition discussed in Section~2.4. On the 2D Two-Room task, the triggered width expansion (seed 42) confirms a representational bottleneck: the encoder required more independent dimensions to resolve the task. On the 60D 30-Object task, the triggered depth expansions (seeds 3072, 123) confirm a computational bottleneck: the encoder had sufficient dimensionality ($d_\text{model}=64$) but lacked the hierarchical depth to model complex multi-object interactions. On Push-T, the absence of expansion indicates that neither bottleneck was active, as the 5D state was fully representable and computable with the minimal architecture.

\paragraph{Efficiency}
Table~\ref{tab:efficiency} shows that SCG effectively bridges the gap between under-provisioned and over-provisioned models. On 30-Object, SCG achieves a loss reduction of 0.0017 using only 50K additional parameters over Fixed Small, whereas Fixed Large requires 5.4M additional parameters to achieve a reduction of 0.0033. This makes SCG approximately 56$\times$ more parameter-efficient in translating capacity growth to predictive improvement.

\paragraph{The over-provisioning penalty.}
A notable observation on the Two-Room task is that the SCG encoder (seed 42) not only surpassed Fixed Small but also outperformed the Fixed Large model in prediction loss (0.000176 vs. 0.000362), despite using only 11\% of its parameters. This indicates an \emph{over-provisioning penalty}: when the model capacity vastly exceeds the task complexity, the excess parameters capture spurious correlations or noise (over-fitting), and the optimizer struggles to navigate a high-dimensional, redundant loss landscape. By starting minimal and growing only a single width dimension, SCG maintains a tighter, more regularized representation space that generalizes better to the simple 2D dynamics than the over-provisioned 12-layer architecture.

\begin{table}[H]
\centering
\caption{Parameter efficiency. SCG closes the performance gap to Fixed Large with up to 56$\times$ greater parameter efficiency than simply scaling the architecture.}
\label{tab:efficiency}
\begin{tabular}{lccc}
\toprule
\textbf{Config} & \textbf{Params} & \textbf{Relative to Fixed Large} & \textbf{Avg $\mathcal{L}_\text{pred}$} \\
\midrule
Fixed Small & 283K & 5.0\% & 0.008368 \\
\textbf{SCG} & \textbf{333K} & \textbf{5.8\%} & \textbf{0.006666} \\
Fixed Large & 5.7M & 100\% & 0.005080 \\
\bottomrule
\end{tabular}
\end{table}

\begin{figure}[H]
\centering
\includegraphics[width=\textwidth]{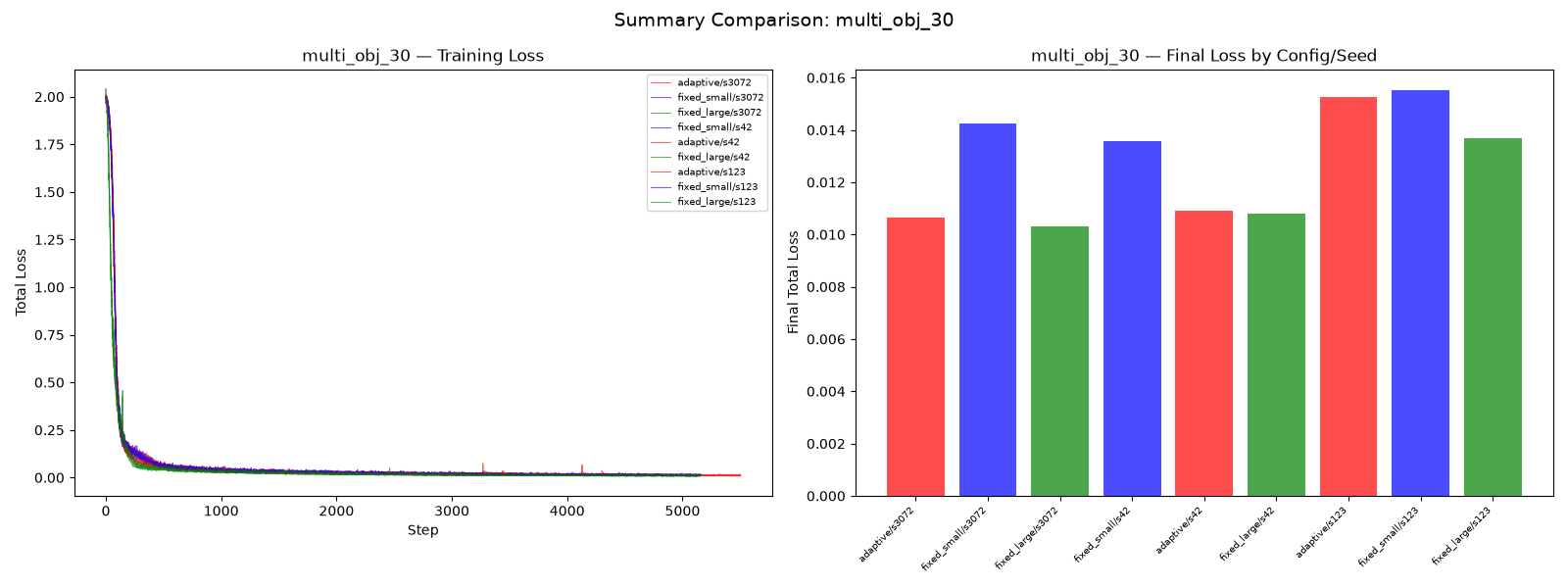}
\caption{30-Object total training loss comparison. SCG (red) achieves lower loss than Fixed Small (blue) via depth expansion, while using far fewer parameters than Fixed Large (green).}
\label{fig:loss_comparison}
\end{figure}

\section{Limitations}

\begin{enumerate}
\item \textbf{Synthetic environments.} All experiments use synthetic environments. The expansion mechanism should be validated on real data with higher visual complexity.

\item \textbf{No planning evaluation.} LeWM's downstream task is planning via Cross-Entropy Method. We evaluate prediction loss but not planning success rate, the ultimate metric for world models.

\item \textbf{Single GPU scale.} Models up to 5.7M parameters were tested. Scaling to ViT-Base (86M) is untested.

\item \textbf{Seed 42 on 30-Object.} One of three seeds on the 30-Object task did not plateau within 50 epochs, preventing expansion. Longer training or a lower plateau threshold would address this, but at the cost of more test epochs.

\end{enumerate}

\section{Discussion \& Outlook}

\textbf{Successive learning.} The core insight is that encoder capacity need not be pre-allocated at maximum - it can grow \emph{successively}, guided by the task's demands. Width expansion provides low-level semantic capacity (more independent representation dimensions), while depth expansion provides higher-order semantic abstraction (hierarchical transformations). SIGReg's complex characteristic function matching ensures that all grown dimensions remain statistically independent and aligned with the predictive objective, preventing the redundancy that plagues fixed architectures.

\textbf{Redundancy in fixed ViT.} Our analysis confirms that LeWM's ViT-Tiny encoder contains significant redundancy: Layers 3--12 contributing progressively less (residual ratios 0.9 $\to$ 0.15). This suggests that the standard practice of pre-allocating large encoders wastes compute on redundant processing. SCG starts minimal and grows only what is needed, closing the performance gap to the fixed large baseline with up to 56$\times$ greater parameter efficiency, proving that over-provisioned fixed architectures waste substantial compute on redundant processing..

\textbf{Compute and data efficiency.} By starting with 283K parameters instead of 5.7M, SCG trains faster per epoch and is less prone to overfitting on small datasets. The test-and-verify mechanism adds at most 4 extra epochs per plateau (2 for width test, 2 for depth test).

\textbf{The over-provisioning penalty.} The fact that SCG surpassed the Fixed Large model on the 2D task challenges the standard practice in deep  learning of using the largest feasible architecture. Our results suggest that over-provisioning is not merely computationally wasteful but can actively degrade predictive performance by introducing redundant parameters that capture noise rather than signal. SCG inherently avoids this penalty by allocating capacity successively, ensuring that the architecture remains tightly coupled to the task's actual dimensionality and complexity.

\textbf{Broader impact.} Beyond mere computational savings, the core significance of Successive Capacity Growth lies in its ability to foster \emph{target-ordered learning}. By growing width only for new independent semantic factors and depth only for higher-order abstractions over existing ones, the architecture is forced to build a well-structured, non-redundant representation space. Unlike fixed, over-provisioned models that often distribute information arbitrarily across redundant heads and layers, SCG ensures stronger that semantic dimensions are added purposefully and build upon each other hierarchically. This architecture-agnostic principle of orderly, demand-driven semantic construction could benefit transformer-based models broadly, shifting the paradigm from "allocate maximum capacity and hope it organizes itself" to "grow structured semantics only when the task demands it."

\textbf{Future directions.} (1) Testing on real data to validate on higher visual complexity. (2) Integrating with the full LeWM planning pipeline to evaluate downstream task success. (3) Pruning mechanisms that remove unused heads or layers when the task simplifies. (4) Combining SCG with dynamic depth and width adjustment mechanisms \citep{zhang2026recursivevisiontransformerdynamic} for further resource efficiency during inference.  (5) Extending to larger architectures (ViT-Base) where the redundancy savings would be more substantial.

\section{Conclusion}

We proposed Successive Capacity Growth (SCG), a method that grows ViT encoders in width (for low-level semantic capacity) or depth (for higher-order semantic abstraction) via function-preserving expansions triggered by a task-agnostic test-and-verify mechanism. SCG starts from 283K parameters and grows only when the task demands, achieving 49\% improvement on a 2D task (width expansion) and 20.3\% on a 60D task (depth expansion) over the fixed small baseline, closing the performance gap to the fixed large baseline with up to 56$\times$ greater parameter efficiency. The key insight is that world model encoders need not be pre-allocated at maximum capacity - they can grow successively as the task demands, with SIGReg's complex characteristic function matching ensuring that all grown dimensions remain statistically independent and non-redundant.

\bibliography{ExampleBibFile}

\appendix

\section{Hyperparameters}
\label{app:hyperparams}

\begin{table}[H]
\centering
\caption{Complete hyperparameter table.}
\label{tab:hyperparams}
\begin{tabular}{ll}
\toprule
\textbf{Parameter} & \textbf{Value} \\
\midrule
Patch size & 14 \\
Image size & 224$\times$224 \\
$d_\text{head}$ & 64 \\
MLP ratio & 4 \\
Projector dim & 192 \\
Action emb dim & 64 \\
Predictor hidden & 256 \\
SIGReg $\lambda$ & 0.1 \\
SIGReg sketch\_dim & 64 \\
SIGReg num\_knots & 17 \\
Learning rate & 3e-4 (500-step warmup) \\
Weight decay & 0.05 \\
Batch size & 128 \\
Gradient clipping & max\_norm=1.0 \\
Plateau threshold & 2\% improvement over 2 epochs ($\mathcal{L}_\text{pred}$) \\
Improvement threshold & 2\% (same as plateau, $\mathcal{L}_\text{pred}$) \\
Test window & 2 epochs \\
Cooldown & 2 epochs \\
Effective epochs & 50 \\
Max heads / layers & 12 / 12 \\
Max params & 15M \\
\bottomrule
\end{tabular}
\end{table}

\section{Reproducibility Checklist}
\label{app:reproducibility}

\begin{itemize}
\item Seeds: 3072, 42, 123 (all experiments)
\item Data generation seed: 42 (fixed across all runs)
\item Hardware: NVIDIA RTX A6000 (48GB VRAM)
\item Framework: PyTorch 2.4.1, CUDA 12.4
\item Parallel execution: 3 configs per (task, seed) via ThreadPoolExecutor
\item Effective epochs: 50 (test epochs for rollbacks do not count)
\item All code available at \url{https://github.com/121-labs/ViT-Expansion-in-JEPA-WM}
\end{itemize}

\end{document}